\documentclass[11pt]{article}
\usepackage{coling2020}
\usepackage{times}
\usepackage{url}
\usepackage{latexsym}
\usepackage{graphicx}
\usepackage{tabularx}
\usepackage{booktabs}
\usepackage{todonotes}
\usepackage{verbatim}
\usepackage{xcolor}
\usepackage{multirow}
\usepackage{subcaption}
\usepackage[multiple]{footmisc}

\colingfinalcopy 

\title{Matching Theory and Data with Personal-ITY: What a Corpus of Italian YouTube Comments Reveals About Personality}

\author{Elisa Bassignana$^{\diamondsuit\heartsuit}$ \quad Malvina Nissim$^\clubsuit$ \quad Viviana Patti$^\heartsuit$\\ 
$^\heartsuit$Dipartimento di Informatica, Università degli Studi di Torino, Italy \\
$^\diamondsuit$Department of Computer Science, IT University of Copenhagen, Denmark\\
$^{\clubsuit}$ CLCG -- Faculty of Arts, University of Groningen, The Netherlands\\
{\tt  $^\heartsuit$viviana.patti@unito.it,
$^{\diamondsuit}$elba@itu.dk}\\ \texttt{$^{\clubsuit}$m.nissim@rug.nl}\\}

\date{}

\begin{document}
\maketitle
\begin{abstract}
As  a  contribution to personality detection in languages  other than English, we rely on distant supervision to create Personal-ITY, a novel corpus of YouTube comments in Italian, where authors are labelled with personality traits. The traits are derived from one of the mainstream personality theories in psychology research, named \textit{MBTI}. Using personality prediction experiments, we (i) study the task of personality prediction in itself on our corpus as well as on {\scshape TwiSty}, a Twitter dataset also annotated with MBTI labels; (ii) carry out an extensive, in-depth analysis of the features used by the classifier, and view them specifically under the light of the original theory that we used to create the corpus in the first place. We observe that no single model is best at personality detection, and that while some traits are easier than others to detect, and also to match back to theory, for other, less frequent traits the picture is much more blurred.

\end{abstract}

\section{Introduction}
\label{intro}
%
%
\blfootnote{
    %
    %
    %
    %
    %
    %
     \hspace{-0.65cm}  
     This work is licensed under a Creative Commons 
     Attribution 4.0 International License.
     License details:
     \url{http://creativecommons.org/licenses/by/4.0/}.
}

Human Personality is a psychological construct aimed at explaining the wide variety of human behaviours in terms of a few, stable and measurable individual characteristics \cite{snyder1983influence,parks2009personality,vinciarelli2014survey}.
Research in psychology has formalised these characteristics into what are known as \textit{Trait Models}. Two are the major models widely adopted also outside of purely psychological research (see Section \ref{sec:psychologicalmodels}): \emph{Big Five} \cite{John1999TheBF} and \emph{Myers-Briggs Type Indicator} (\emph{MBTI}) \cite{myers1995gifts}.


The psychological tests commonly used to detect prevalence of traits include human judgements regarding semantic similarity and relations between adjectives that people use to describe themselves and others. This is because language is believed to be a prime carrier of personality traits \cite{schwartz2013personality}. This aspect, together with the progressive increase of available user-generated data from social media, has prompted the task of \emph{Personality Detection}, i.e., the automatic prediction of personality from written texts
\cite{whelan2006profiling,automatprof,celli2013workshop,youyou2015computer,litvinova,Verhoeven2016TwiStyAM}. 
Personality detection can be useful in predicting life outcomes such as substance use, political attitudes and physical health. Other fields of application are marketing, politics, psychological and social assessment and, in the computational domain, dialogue systems \cite{Ma2020ASO} and chatbots \cite{ijcai2018-595}.

As a contribution to personality detection in languages other than English, we have developed Personal-ITY, a novel corpus of YouTube comments in Italian, which are annotated with MBTI personality traits.
The corpus creation methodology, described in detail in \cite{bassignanaClicIt}, makes use of a Distant Supervision approach that can also serve as a blueprint to develop datasets for other languages. 
In this work, we use Personal-ITY to cast light not only on the feasibility of personality detection {\em per se} on this corpus, but also on the relationship between our data and the theory it is based on. More specifically, we want to investigate if our distantly obtained labels are meaningful with respect to the psychological theory they come from, and whether language does indeed reflect the traits that should be associated with such labels. To do this, we run a series of in- and cross-dataset experiments; on top of performance analysis we conduct an in depth study on the relevant features used by the classifiers, and how they might relate to the source psychological theory.

Personal-ITY is available at \url{https://github.com/elisabassignana/Personal-ITY}.

\section{Background}

Personality profiling is  addressed both from a psychological viewpoint (traits model for the classification of personality) and from a computational perspective in the field of Natural Language Processing. We provide relevant background from both sides, as they are intertwined in our work.

\subsection{Psychological models}
\label{sec:psychologicalmodels}
There are two main personality trait models widely accepted and used by the research community, also outside of psychology: \textit{Big Five} and \textit{Myers-Briggs Type Indicator} (\textit{MBTI}).

Big Five \cite{John1999TheBF}, also know as Five-Factor Model (FFM) or the OCEAN model, was developed from the 1980s onwards. This theory outlines five global dimensions of personality and describes people by assigning a score in a 
range for each of them.
The five traits considered are: {\scshape Openness to experience}, {\scshape Conscientiousness}, {\scshape Extroversion}, {\scshape Agreeableness} and {\scshape Neuroticism}. Thus a person's personality would be defined through five corresponding scores indicating the positive or negative degree to which each dimension is expressed. Interestingly, the Big Five tests usually assign scores on the basis of semantic associations between personality traits and words considering the texts of the users' answers,
rather than relying on neuropsychological experiments.
The use of the Big Five model in computational approaches to personality began more than one decade ago and it is now widely accepted in academia.

The MBTI \cite{myers1995gifts} theory is based on the conceptual theory of the Psychological Types proposed by the psychiatrist Carl Jung, who had speculated the main dimensions able to describe how people experience the world. Assessment is based  on a self-reported psychological questionnaire that helps researchers to classify people into one personality type out of sixteen. The sixteen labels are the product of binary labels over four different dimensions, as follows: ({\scshape Extravert-Introvert}, {\scshape iNtuitive-Sensing}, {\scshape Feeling-Thinking}, {\scshape Perceiving-Judging}). Each person is assumed to have one dominant quality from each category, thus producing sixteen unique types. Examples of full personality types are therefore four letter labels such as {\scshape ENTJ} or {\scshape ISFP}.

The initial intention of the test was to help women who were entering the industrial workforce for the first time 
during the Second World War 
to identify the ``best, most comfortable and effective" job for them based on their personality type. In later years, MBTI continued to be used in order to predict validity of employees' job performance and to help students in their choice of career or course of study.

Although several studies suggest that the MBTI test lacks convincing validity data for these types of applications as it can measure preferences and not ability, it continues to be popular because it is very easy to administer it and it is not difficult to understand. 

\subsection{Personality Detection}
\label{sec:back}

Most approaches to automatic personality detection are supervised models trained on silver or gold labelled data. In this section, we revise existing datasets and standard methods of personality detection.

\paragraph{Corpora} There exist a few datasets annotated for personality traits. For the shared tasks organised within the \emph{Workshop on Computational Personality Recognition} 
\cite{celli2013workshop}, four English datasets annotated with the \textit{Big Five} traits have been released.
For the 2013 edition, the data contained ``Essays" \cite{essays}, which is a large dataset of stream-of-consciousness texts collected between 1997 and 2004, and ``myPersonality"\footnote{\url{http://mypersonality.org}}, a corpus collected from FaceBook including information on user social network structures.
For the 2014 edition, the data consisted of the ``YouTube Personality Dataset" \cite{biel2013youtube}, which contains a collection of behavioural features, speech transcriptions, and personality impression scores for a set of 404 YouTube vloggers, and ``Mobile Phones", which is a collection of call logs and proximity data of 53 subjects living in a student residency of a major US university, collected through a special software incorporated in their phones
\cite{mobileDataset}.

\newcite{schwartz2013personality} collected a Big Five annotated dataset of FaceBook comments (700~millions~words) written by 136.000 users who shared their status updates. Interesting correlations were observed between word usage and personality traits.

For the 2015 PAN Author Profiling Shared Task \cite{Pardo2015OverviewOT}, personality was added to gender and age in their standard profiling task, with tweets in English, Spanish, Italian and Dutch annotated according to the \textit{Big Five} model assigning a score in a range [-0.5; +0.5] for each trait.


If looking at data labelled with the MBTI traits, we find a corpus of 1.2M English tweets annotated with personality and gender \cite{Personality_Traits_on_Twitter}, and the multilingual dataset
{\scshape TwiSty} \cite{Verhoeven2016TwiStyAM}.
The latter is a corpus of data collected from Twitter using a Distant Supervision approach.
It is annotated with MBTI personality labels and gender for six languages (Dutch, German, French, Italian, Portuguese and Spanish), and includes a total of  18,168 authors. 

As we concentrate on Italian, we report in Table \ref{tab:corpus} an overview of the available Italian corpora labelled with personality traits. We include information on our own Personal-ITY corpus, which is described in Section~\ref{sec:corpus}. For {\scshape TwiSty}, we only report information for the Italian portion. 

\begin{table}
\centering
\begin{tabular}{l|lrr}
\toprule
Corpus & Model & \# user & Avg. \\
\midrule
PAN2015 & Big Five & 38 & 1,258 \\
{\scshape TwiSty} & MBTI & 490 & 21,343 \\
Personal-ITY & MBTI & 1048 & 10,585 \\
\bottomrule
\end{tabular}
\caption{Summary of Italian corpora  with personality labels. Avg.: average tokens per user.}
\label{tab:corpus}
\end{table}

Computational work exists also on comparing (labels from) the two models \cite{Celli2018IsBF}. \newcite{furnham} defined some correlations between various dimensions across the two trait models: Big Five Extraversion is correlated with MBTI Extraversion-Introversion, Openness to Experience is correlated with Sensing-Intuition, Agreeableness with Thinking-Feeling and Conscientiousness with Judging-Perceiving. In order to increase the amount of data we could work with, and to obtain a general, usable model, we tried to convert the PAN~2015 Italian data to MBTI annotations. We considered the mid value in the Big Five range as threshold between the opposite poles of MBTI dimensions.
This experiments didn't led to any informative results, probably due the the small dimension of the corpus (see Table \ref{tab:corpus}): almost all the few users present have been annotated with the same MBTI label.

\paragraph{Detection Approaches} Regarding detection approaches, 
\newcite{mairasse} tested the usefulness of different sets of textual features making use of mostly SVMs.
At the PAN~2015~challenge (see above) a variety of algorithms were tested (such as Random Forests, decision trees, logistic regression for classification, and also various regression models), but overall most successful participants used SVMs. Regarding features, 
participants approached the task with combinations of style-based and content-based features, as well as their combination in \emph{n}-gram models \cite{Pardo2015OverviewOT}.

Experiments on {\scshape TwiSty} were performed by the corpus creators themselves using a LinearSVM with  word (1-2) and character (3-4) \emph{n}-grams. Their results (reported in Table~\ref{expTSnostri1} for the Italian portion of the dataset) are obtained through 10-fold cross-validation; the model is compared to a weighted random baseline (WRB) and a majority baseline (MAJ).
Let us notice that the model proposed in \cite{Verhoeven2016TwiStyAM} 
for the Italian language is the only one
not reaching any baseline (for all the other languages the model proposed reach at least the
weighted random baseline).
This also prompted us to work on the Italian language, where there is still ample room for improvement on the development of resources and models for the personality detection task.
More in general, our choice has to be seen as an intention of improving the state of the art for languages other than English \cite{joshi-etal-2020-state}.

Recent computational approaches on personality prediction from texts investigated the use of deep learning \cite{navonil} and regression models \cite{akrami2019automatic}.

\section{Data}
\label{sec:corpus}

To run experiments on personality detection, we have created a dedicated corpus with MBTI labels, exploiting distant supervision: Personal-ITY \cite{bassignanaClicIt}. Here, we summarise the choices that we made regarding the source of the data and the theoretical trait model, the procedure followed to construct the corpus, and provide a description of the resulting dataset. In addition, we also partly use the existing {\scshape TwiSty} (see Section~\ref{sec:back} and Table~\ref{tab:corpus}).

Because we deal with personal data, and because we do believe profiling is a sensitive task in general, we also provide an Ethics Statement.

\subsection*{Ethics Statement}

Personality profiling must be carefully evaluated from an ethical point of view. In particular personality detection can involve ethical issues regarding the appropriate use and interpretation of the prediction outcomes \cite{ethicalconsiderations}. Also, concerns have been raised regarding the inappropriate use of these tests with respect to invasion of privacy, cultural bias and confidentiality \cite{mehta2019recent}.

The data included in the Personal-ITY dataset was publicly available on the YouTube platform at the time of the collection. As we explain in this Section (but see \cite{bassignanaClicIt} for details), the information collected consists in comments published under public videos on the YouTube platform by the authors themselves.
For an increased protection of user identities, in the released corpus only the YouTube usernames of the authors are mentioned, which are not unique identifiers.
The YouTube IDs of the corresponding channels, which are instead unique identifiers on the platform and would allow to trace back the identity of the authors, are not released.
The corpus was created for academic research purposes, and is not intended for commercial deployment or applications.

\subsection{Source and Theoretical Model}
 YouTube is the source of data for our corpus. The decision is grounded on the fact that compared to the more commonly collected tweets, YouTube comments can be longer, so that 
users are freer to express themselves without limitations. Additionally, there is a substantial amount of available data on the YouTube platform, which is easy to access thanks to the free YouTube APIs. 

Our theoretical trait model of choice is MBTI. 
Although it has been extensively criticized for a number of limitations \cite{McCrae,Boyle,Pittenger}, and the Big Five model seems to be more widely accepted in psychology, our choice has been driven by two main reasons.
The first benefit of this decision is that MBTI is easy to use in association with a Distant Supervision approach (just checking if a message contains one of the 16 personality types; see Section~\ref{sec:ds}).
Another benefit is related to the existence of {\scshape TwiSty}. Since both {\scshape TwiSty} and Personal-ITY implement the MBTI model, analyses and experiments over personality detection can be carried out also in a cross-domain setting. 

\subsection{Corpus Creation and Description}
\label{sec:ds}

The fact that users often self-disclose information about themselves on social media makes it possible to adopt \textit{Distant Supervision} (DS) for the acquisition of training data. DS is a semi-supervised method that has been abundantly and successfully used in affective computing and profiling to assign silver labels to data on the basis of indicative proxies \cite{go2009twitter,pool2016distant,emmery-etal-2017-simple}.

We observed that some YouTube Italian users were used to leave comments to videos on the MBTI theory, in which they were stating their own personality type
(e.g. \emph{Sono ENTJ...chi altro?} [en: ``I'm ENTJ...anyone else?"]; \emph{INTP, primo test che effettivamente ha ragione} [en: "INTP, the first test that is actually right"]). We exploited such comments to create Personal-ITY. 

The methodology, explained in detail in \cite{bassignanaClicIt}, consisted in creating automatically a list of YouTube users annotated with MBTI personality labels starting from the comments cited above. In the second macro-step, we adopted a Distant Supervision approach in order to retrieve as much as possible texts written by the authors whose personality was known.
The result of this procedure led to a final corpus with a conspicuous number of users and comments, where only  authors with at least five comments, each at least five token long, are included.

Personal-ITY contains 1048 users, each annotated with an MBTI label. The average number of comments per user is 92 and each message is on average 115 tokens.
Table \ref{statistics} shows explicitly the comparison between our new corpus and {\scshape TwiSty}.

\begin{table}[h]
\centering
\begin{tabular}{l|rrrr}
\toprule
Corpus & \# Users & Avg. comments/user & Avg. tokens/comment & Avg. tokens/user \\
\midrule
Personal-ITY & 1048 & 92 & 115 & 10,585 \\
{\scshape TwiSty} & 490 & 1,903 & 11 & 21,343 \\
\bottomrule
\end{tabular}
\caption{Statistical comparison between Personal-ITY and {\scshape TwiSty}.}
\label{statistics}
\end{table}

\begin{figure}
\centering
\includegraphics[width=0.7\textwidth]{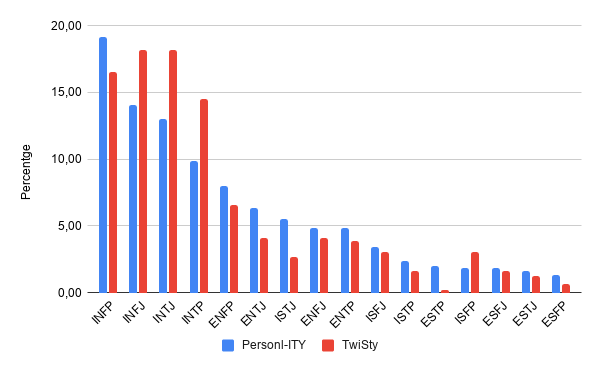}
\caption{Distribution of the 16 labels in the YouTube corpus and in the Italian part of {\scshape TwiSty}.}
\label{fig:percUsers}
\end{figure}

The amount of the 16 personality types in the corpus is not uniform. Figure \ref{fig:percUsers} shows such distribution and also compares it with the one in {\scshape TwiSty}.
It can be observed that the two corpora present quite similar percentages of personality types. 
Some little differences (e.g., \emph{INFJ}, \emph{INTJ}, \emph{INTP}) can be explained considering the different data sources: Personal-ITY is collected from YouTube, while {\scshape TwiSty} is collected from Twitter.
The general unbalanced distribution can be due to personality types not being uniformly distributed in the population, and to the fact that different personality types can make different choices about their online presence.
\newcite{PersonalityandonlineofflinechoicesMbtiprofiles} for example, observed that there is a significant correlation between online--offline choices and the MBTI dimension of {\scshape Extravert-Introvert}: 
extroverts are more likely to opt for offline modes of communication, while online communication is presumably easier for introverts. Figure~\ref{fig:percUsers} confirms this theory as the four most frequent types are introverts in both datasets.
The conclusion is that, despite the different biases, collecting linguistic data in this way has the advantages that it reflects actual language use and allows large-scale analysis \cite{Personality_Traits_on_Twitter}.

\section{Experiments}
\label{experiments}

We ran a series of initial experiments on Personal-ITY, which we use to peek into the relationship between labels and theory, and which can also serve as a baseline for future work on this dataset.

The choices related to the experimental framework have been driven by the interest on investigating signals and linguistic cues for personality traits, analyzed through the lens of psychological studies on personality. This perspective has therefore led us not use deep learning techniques that would have made this analysis task more complex. Rather, we opt for state-of-the-art approaches commonly used in author profiling, developing interpretable models that can aid the analysis process. These will make it possible to perform a linguistic and psychological analysis, though perhaps at the expense of performance.

Specifically, we used the \verb|sklearn| \cite{scikit-learn} implementation of a linear~SVM (LinearSVM), with standard parameters, and tested three types of features: lexical-, stylistic-, and embeddings-based. We used four placeholders for hashtags, urls, usernames and emojis. 

At the lexical level, we experimented with word (1-2) and character (3-4) \emph{n}-grams, both as raw counts as well as tf-idf weighted. 
Character \emph{n}-grams were tested also with a word-boundary option.
Considering stylistic features, we investigated the use of emojis, hashtags, pronouns, punctuation and capitalisation.
Lastly, we also experimented with embeddings-based representations, using more generic \cite{embGloVe} and YouTube-specific \cite{embYT} pre-trained models. We  created one representation per user by averaging the vectors for all words written by that user.

We used 10-fold cross-validation, and assessed the models using macro f-score. We deem this way of averaging over f-scores per class appropriate, since the dataset is quite unbalanced, but we want good performance for each class. For comparison, we calculated a majority baseline (MAJ).

Table~\ref{expYT} shows the results of our experiments with the different feature types described above. 
Regarding \emph{n}-grams and embeddings representations, we report results for the best configurations, namely character \emph{n}-grams for the lexical features, and GloVe embeddings.
Overall, lexical features perform best. Combining different feature types did not lead to any improvement. Classification was performed both with four separate binary classifiers (one per dimension), as well as with one single classifier predicting a total of four classes, i.e, the whole MBTI labels at once. Interestingly, in the latter case, we observe that the results are quite high considering the increased difficulty of the task.

Table~\ref{expTSnostri} reports the scores of our models on {\scshape TwiSty}. Because the original {\scshape TwiSty} paper uses micro f-score, for the sake of comparison, in Table~\ref{expTSnostri1} we include the results of our experiments using micro-f for the MAJ baseline and our lexical \emph{n}-gram model. For all traits, our models achieve better results (micro-f) than those reported in the original {\scshape TwiSty} paper \cite{Verhoeven2016TwiStyAM}. In Table~\ref{expTSnostri2}, instead, there are the macro-f of the experiments we performed on {\scshape TwiSty}.
As for Personal-ITY, best results were achieved using lexical features (tf-idf word \emph{n}-grams); results of models with stylistic features and embeddings were just above the baseline.

\begin{table}[h]
\centering
\begin{tabular}{l|cccc|c}
\toprule
\multirow{2}{*}{Trait} & \multicolumn{4}{c|}{Binary classification} & Full label at once \\
\cline{2-6}
 & MAJ & \emph{n}-grams & Sty & Emb & \emph{n}-grams \\
\toprule
EI & 40.55 & 51.85 & 40.46 & 40.55 & 51.65 \\
NS & 44.34 & 51.92 & 44.34 & 44.34 & 49.04 \\
FT & 35.01 & 50.67 & 36.27 & 35.01 & 50.86 \\
PJ & 29.49 & 50.53 & 51.04 & 47.06 & 51.03 \\
\midrule
Avg & 37.35 & \textbf{51.24} & 43.03 & 41.74 & 50.65 \\
\bottomrule
\end{tabular}
\caption{Results of the experiments on Personal-ITY. Predictions of the full MBTI label at once were performed using the model performing best in the binary classification.}
\label{expYT}
\end{table}

\begin{table}[h]
\begin{minipage}{0.6\textwidth}
\begin{subtable}[h]{0.87\textwidth}
\begin{tabularx}{\textwidth}{l|ccc|cc}
\toprule
& \multicolumn{3}{c|}{\cite{Verhoeven2016TwiStyAM}} & \multicolumn{2}{c}{Our experiments} \\
\midrule
Trait & WRB & MAJ & Lex & MAJ & \emph{n}-grams \\
\toprule
EI & 65.54 & 77.88 & 77.78 & 77.75 & \textbf{79.18} \\
NS & 75.60 & 85.78 & 79.21 & 85.92 & \textbf{85.92} \\
FT & 50.31 & 53.95 & 52.13 & 53.67 & \textbf{55.31} \\
PJ & 50.19 & 53.05 & 47.01 & 53.06 & \textbf{54.08} \\
\midrule
Avg & 60.41 & 67.67 & 64.06 & 67.6 & \textbf{68.62} \\
\bottomrule
\end{tabularx}
\subcaption{Comparison between  our experiments on {\scshape TwiSty} and the ones in \cite{Verhoeven2016TwiStyAM}. Scores reported are micro f-scores. WRB = weighted random baseline. MAJ = majority baseline.}
\label{expTSnostri1}
\end{subtable} 
\end{minipage}
\begin{subtable}[h]{0.4\textwidth}
\begin{tabular}{l|cccc}
\toprule
Trait & MAJ & \emph{n}-grams & Sty & Emb \\
\toprule
EI & 43.69 & 55.23 & 43.69 & 43.69 \\
NS & 46.15 & 46.15 & 46.15 & 46.15 \\
FT & 34.79 & 52.98 & 35.34 & 34.70 \\
PJ & 34.56 & 53.01 & 35.20 & 34.90 \\
\midrule
Avg & 39.80 & \textbf{51.84} & 40.09 & 39.86 \\
\bottomrule
\end{tabular}
\caption{Results of our experiments on {\scshape TwiSty} by using macro f-scores.}
\label{expTSnostri2}
\end{subtable}
\caption{Results of our experiments on {\scshape TwiSty}.}
\label{expTSnostri}
\end{table}

To test compatibility of resources and to assess model portability, we also ran cross-domain experiments on Personal-ITY and {\scshape TwiSty}. We divided both corpora in fixed training and test sets with a proportion of 80/20 so that the test set stays the same for in-domain and cross-domain settings. Indeed, we run the in-domain models again using this split. The models use lexical features as they are the one performing better overall the others. Results are shown in Table~\ref{expCrossDomain}.
We ran the experiments both using a binary classification of each trait separately and with the prediction of the full MBTI label at once.
This latter leads to results even better on the considered sets.
Cross-domain scores are obtained with the best in-domain model\footnote{Binary classification: Train on Personal-ITY: character \emph{n}-grams. Train on {\scshape TwiSty}: tf-idf character \emph{n}-grams.}\footnote{Full MBTI label at once: Train on Personal-ITY: character \emph{n}-grams. Train on {\scshape TwiSty}: tf-idf word \emph{n}-grams.}.
They drop substantially compared to in-domain, but are always above the baseline.

We specify that in every experiments done (Tables~\ref{expYT}--\ref{expTSnostri}--\ref{expCrossDomain}) we chose to look for the best model able to predict the whole MBTI personality and so we report the highest scores based on averages of the four traits. Considering the four dimensions individually better results can be obtained by using specific models.
Table~\ref{analysisBestModel} shows an overview of the best models considering each trait independently for each source of data we tested.
Results are quite scattered: 
there is no a single best model for personality predictions, as feature contribution depends on the dimension considered, and on the dataset.
This observation confirms the inherent difficulty of the Personality Detection task from written texts.

\begin{table}[h]
\begin{tabularx}{\textwidth}{X|X|XX|X|XX||X|XX|X|XX}
\toprule
& \multicolumn{6}{c||}{Binary classification} & \multicolumn{6}{c}{Full MBTI label at once}\\
\midrule
Train & \multicolumn{3}{c|}{Personal-ITY} & \multicolumn{3}{c||}{\scshape TwiSty} & \multicolumn{3}{c|}{Personal-ITY} & \multicolumn{3}{c}{\scshape TwiSty} \\
\midrule
\multirow{2}{*}{Test} & \multicolumn{1}{c|}{IN} &  \multicolumn{2}{c|}{CROSS} & \multicolumn{1}{c|}{IN} & \multicolumn{2}{c||}{CROSS} & \multicolumn{1}{c|}{IN} &  \multicolumn{2}{c|}{CROSS} & \multicolumn{1}{c|}{IN} & \multicolumn{2}{c}{CROSS} \\
\cline{2-13}
&  Pers & MAJ  & {\scshape Twi} & {\scshape Twi}  &  MAJ &  Pers &  Pers & MAJ  & {\scshape Twi} & {\scshape Twi}  &  MAJ &  Pers \\
\toprule
EI & 58.94 & 44.94 & 49.33 & 55.66 & 44.59 & 44.59 & 54.67 & 44.94 & 44.94 & 59.77 & 44.59 & 44.59 \\
NS & 52.88 & 47.87 & 47.31 & 47.87 & 45.31 & 45.31& 48.65 & 47.87 & 47.59 & 47.87 & 45.31 & 45.31 \\
FT & 49.20 & 37.58 & 47.09 & 65.26 & 39.13 & 51.04 & 54.18 & 37.58 & 46.38 & 61.98 & 26.32 & 46.71 \\
PJ & 54.43 & 32.41 & 32.50 & 56.87 & 36.56 & 38.54 & 58.07 & 32.41 & 38.08 & 50.27 & 36.56 & 46.80 \\
\midrule
Avg & 53.86 & 40.70 & 44.06 & 56.42 & 41.40 & 44.87 & 53.89 & 40.70 & \textbf{44.25} & 54.97 & 38.20 & \textbf{45.85} \\
\bottomrule
\end{tabularx}
\caption{Results of the cross-domain experiments. MAJ = baseline on the cross-domain testset.}
\label{expCrossDomain}
\end{table}

\begin{table}[h]
\small
\centering
\begin{tabular}{l|c|c|c|c|c|c|c}
\toprule
\multirow{2}{*}{Source} & \multirow{2}{*}{Trait} & \multicolumn{3}{c|}{Tf-idf} & \multicolumn{3}{c}{Count} \\
& & word & char & char\_wb & word & char & char\_wb \\
\midrule
\multirow{4}{*}{Personal-ITY} & EI & & & & & & X \\
 & NS & & & & & & X \\
 & FT & & & X & & & \\
 & PJ & & X & & & & \\
\midrule
\multirow{4}{*}{{\scshape TwiSty}} & EI & X & & & & &  \\
 & NS & & & & & X & \\
 & FT & & & X & & & \\
 & PJ & X & & & & & \\
\midrule
Cross-domain: & EI & & & & X & &  \\
Tr: Personal-ITY & NS & & & X & & & \\
Te: {\scshape TwiSty} & FT & & & & & & X \\
 & PJ & & & & & & X \\
\midrule
Cross-domain: & EI & & & & & X & \\
Tr: {\scshape TwiSty} & NS & & & & & X & \\
Te: Personal-ITY & FT & X & & & & & \\
 & PJ & & & & & & X \\
\bottomrule
\end{tabular}
\caption{Best model for each trait considering each source individually.} 
\label{analysisBestModel}
\end{table}

The in-domain experiments show that performance over the two datasets is very similar overall, though with some differences regarding the best and worst predicted traits. However, we observe that cross-domain performance drops by approximately 10 points, independently of the direction of training and testing taken. This underlines differences in the dataset which might not make them fully compatible. For our in-depth linguistic analysis, we choose to concentrate on Personal-ITY, mainly due to the availability of longer author's comments, which can give rise to more interesting insights when studying word-based feature contribution in connection with the source MBTI personality theory.

\section{Feature Analysis and Linguistic Cues for Personality Traits}
\label{analysis}

In this section we discuss possible correlations between linguistic cues derived from the experiments described in Section~\ref{experiments} on the Personal-ITY corpus and psychological traits descriptions deriving from the field of Psychology. 
Table \ref{mostInformativefeatures} shows the most important features on which our classifier bases its decisions (i.e., the most relevant ones based on the weight they have on the model prediction).
We report word \emph{n}-gram features 
as they are the most interpretable for a psychological analysis.

\begin{table}[h]
\centering
\small
\begin{tabularx}{\textwidth}{l|ll|XX|Xl|Xl}
\toprule
&Extravert & Introvert & Sensing & Intuition & Thinking & Feeling & Judging & Perceiving \\
\midrule
\multirow{10}{*}{\rotatebox{90}{Count word (1,2)}} &punti & sbagliato & adoro & fuori & \textbf{ma io} & \textbf{io sono} & \textbf{alle} & \textbf{bene} \\
&poi & \textbf{che era} & \textbf{gli} & \textbf{ehm} & nel & giorno & \textbf{un video} & 12\\
&ora & fa & \textbf{del} & sbagliato & \textbf{vero} & beh & \textbf{nella} & \textbf{emoji ho}\\
&ne & via & oddio & \textbf{vorrei} & anni & marco & ancora & che ti\\
&ancora & \textbf{anzi} & credo & morto & \textbf{vedo} & nero & \textbf{tuoi} & non ci\\
&\textbf{emoji non} & hashtag se & qualcuno & che tu & 17 & sotto & \textbf{minecraft} & ho fatto\\
&volevo & \textbf{lui} & solo & alcune & tanto & più & so se & \textbf{beh}\\
&midna & sia & io mi & serie & con le & avevo & \textbf{te} & \textbf{nether}\\
&dal & \textbf{era} & idea & tutto il & chi & \textbf{dentro} & \textbf{metti} & tanto\\
&\textbf{fosse} & \textbf{penso} & va & secondo & \textbf{capire} & \textbf{trovo} & \textbf{neanche} & test\\
\midrule
\multirow{10}{*}{\rotatebox{90}{Tf-idf word (1,2)}}&in & nn & matteo & die & \textbf{perchè} & più & \textbf{user} & perchè\\
&\textbf{xd} & eren & ahah & die die & emoji & emoji emoji & nn & \textbf{emoji}\\
&che & bella playerinside & \textbf{del} & raiden & \textbf{vedo} & marco & \textbf{minecraft} & test\\
&ancora & genio genio & adoro & cane & lullaby & nn & \textbf{da} & \textbf{anche}\\
&ci & playerinside xd & \textbf{di} & cuticole & dei & avevo & \textbf{tuoi} & genio genio\\
&davvero & \textbf{marco} & erenblaze & anche & xd & test & hashtag & non\\
&molto & die & libro & tano & un & genio genio & \textbf{alle} & u3000\\
&di & tifo & marco & \textbf{ehm} & \textbf{questa} & raiden & di eren & \textbf{bene}\\
&perchè & bella & \textbf{in} & copia & eren & in & leo & u3000 u3000\\
&dei & pixelmon & persone & 00 00 & grazie & u3000 & puoi & un\\
\bottomrule
\end{tabularx}
\caption{Most predictive word \emph{n}-grams on Personal-ITY.}
\label{mostInformativefeatures}
\end{table}

Below we are going to analyze each MBTI trait independently by interpreting observed features with existing theoretical definitions of the personality types, which we take from  \cite{geyer2014}.
\begin{itemize}
    \item {\scshape Extravert} vs. {\scshape Introvert} trait:
    \begin{itemize}
        \item {\scshape Extravert}:
        extraverts tend to use abstract words, to be vague and to use adjectives. They talk a lot (in respect to their opposite) about family, friends, groups and social activities.
        \item {\scshape Introvert}:
        introverts, on the other side, tend to use concrete words, to be more precise and so to use nouns, pronouns, articles, numbers and distinctions (\emph{but}, \emph{except}...).
    \end{itemize}
Referring to Table \ref{mostInformativefeatures}, we find \emph{emoji} in the extraversion pole (recall that we normalized all the emojis in the corpus\footnote{\url{https://pypi.org/project/emojis/}}), while it is not present in the opposite pole.
Looking at the definition, we linked this to a more extroverted behavior, people talking about friends, groups and social activities.
Similar to that label there is, in the same column, \emph{xd}, that we intended as an emoticon.
Moreover, closely related to the definition (being abstract and vague) there is the word \emph{fosse} (subjunctive form of the verb `to be').
In the introversion column, instead, there are more concrete and precise words as \emph{che era} (`which was'), \emph{era} (`it was') and \emph{penso} (`I think').
With the same intention to be concrete and precise, those people use (proper) nouns and pronouns as \emph{lui} (`he') and \emph{marco}. 
Lastly, coherently with the introvert definition, we find \emph{anzi} (`rather', `instead'), word belonging to the distinction set.
    
    \item {\scshape Sensing} vs. {\scshape iNtuition} trait:
    \begin{itemize}
        \item {\scshape Sensing}: 
        these people are realistic, they usually talk about practical activities and about what is already happened. They describe facts specifying a lot of details, tangible information and rely a lot on senses.
        \item {\scshape iNtuition}:
        people with this personality follow intuition, fantasy, imagination and ideas. They usually talk about what is going to happen, future possibilities and relate their discourses to abstract and general principles.
    \end{itemize}
In the sensing list of words we highlighted \emph{gli} (`the'), \emph{del} (`of the'), \emph{di} (`of') and \emph{in} (`in') as those tokens are used to specify details. 
In the opposite column, in line with the definition, we just found the words \emph{ehm} and \emph{vorrei} (`I would like to').
    
    \item {\scshape Thinking} vs. {\scshape Feeling} trait:
    \begin{itemize}
        \item {\scshape Thinking}:
        these people follow logic, objectivity, rationality, causality and consistency.
        \item {\scshape Feeling}:
        their opposite, instead, are more inclined to follow the heart and principles; they look for cooperation, harmony and are more sensitive.
    \end{itemize}
In line with the definition of thinking we highlighted the \emph{n}-grams \emph{ma io} (`but I'), \emph{vero} (`true'), \emph{vedo} (`I see'), \emph{capire} (`to understand'), \emph{perchè} (`because') and \emph{questa} (`this').
For their opposite we found \emph{io sono} (`I am'), \emph{dentro} (`inside') and \emph{trovo} (`I think', `I find').    
    
    \item {\scshape Judging} vs. {\scshape Perceiving} trait:
    \begin{itemize}
        \item {\scshape Judging}: 
        this trait indicates determined people, who are used to plan everything and that are comfortable with rules and guide lines.
        \item {\scshape Perceiving}: 
        the opposite are people who like improvisation and tend to keep open options. They are more relaxed and look for liberty.
    \end{itemize}
In the judging column there are words such as \emph{alle} (`at'), \emph{nella} (`in'), \emph{tuoi} (`yours'), \emph{te} (`you'), \emph{metti} (`put'), \emph{neanche} (`neither'), \emph{da} (`from', `by') and \emph{user} (which is a normalized label deriving from pre-processing; it replaces references to specific users). These can all be used to make precise plans, and fit with the description of determined people, comfortable with rules and guidelines.
The more interesting word \emph{n}-grams in this set, is \emph{minecraft}, which is the name of a video game where planning skills are fundamental.
In the opposite pole, perceiving, some distinctive words are: \emph{bene} (`good'), \emph{beh}, the label \emph{emoji} and \emph{anche} (`also'). 

\end{itemize}

A final consideration about the analysis above is that the correlations we found are sometime weak and not so explicit,
especially for the {\scshape S-N} trait and we observed that this is coherent with the not so high results obtained from experiments in Section~\ref{experiments}: it is likely that the absence of strong evidences linking linguistic cues with psychological ones, makes the decision of the classifier hard. 
%
Notice that a similar observation concerning the difficulty to predict the {\scshape S-N} trait has been reported in \cite{Personality_Traits_on_Twitter} on an English corpus, suggesting that such trait could in general be more related to perception, with a weak linguistic signal.
A second observation is that Table~\ref{mostInformativefeatures} contains many unexpected tokens which apparently 
have no explanation.
Some examples are: \emph{midna} for extravert, \emph{eren}, \emph{die} \emph{playerinside} and \emph{pixelmon} for introvert, \emph{erenblaze} for sensing, \emph{die} again also for intuition, \emph{eren}, \emph{17} and \emph{lullaby} for thinking, \emph{raiden} for feeling, \emph{eren} again for judging and \emph{u3000} for perceiving.
The explanation we give to the presence of such `specific' tokens is to be found in the source and in the way we collected the corpus.
Channels on YouTube gather a community of users with similar interests and common saying.
It is therefore likely that users 
commenting videos from a given channel develop some sort of shared, own slang.
In this perspective,  a further analysis of the dataset aimed at detecting topics and word distribution would be useful to identify the presence of a narrow set of specific-domain texts.

For a second qualitative analysis we used \emph{Wordify}\footnote{\url{https://wordify.unibocconi.it/index}}, a tool developed by Bocconi University whose intent is to identify words that discriminate categories in textual data.
Table \ref{wordify} reports the results of such tool on Personal-ITY, and we can observe various correspondences with  Table~\ref{mostInformativefeatures} (terms appearing in both Tables) such as, regarding the first trait, \emph{xd} and \emph{davvero} for {\scshape Extravert} and \emph{eren}, \emph{playerinside xd} and \emph{nn} for {\scshape Introvert}.
This `double check' makes bold words in Table \ref{wordify} reliable features in MBTI personality prediction, even if for most of them we didn't find a direct psychological explanation coherent with the trait definitions.

\begin{table}[h]
\centering
\begin{tabular}{l|lll|l|lll}
\toprule
Trait & Term & Score & Label & Trait & Term & Score & Label\\
\toprule
\multirow{8}{*}{\rotatebox{90}{First trait}} &
sì & 0,384 & E & \multirow{8}{*}{\rotatebox{90}{Second trait}} & emoji & 0,312 & N \\
 & \textbf{xd} & 0,352 & E & \\
& \textbf{davvero} & 0,328 & E & \\
&\textbf{eren} & 0,474 & I  &\\
&emoji & 0,466 & I &\\
&bello \textbf{playerinside} & 0,400 & I & \\
&\textbf{playerinside xd} & 0,324 & I &\\
&\textbf{nn} & 0,312& I &\\
\midrule
\multirow{7}{*}{\rotatebox{90}{Third trait}} &\textbf{test} & 0,414 & F &\multirow{7}{*}{\rotatebox{90}{Fourth trait}} & \textbf{hashtag} & 0,344 & J\\
&qi & 0,364 & F & & \textbf{nn} & 0,336 & J \\
&commento & 0,338 & F & & \textbf{eren} & 0,306 &J \\
&ahah & 0,420 & T & & \textbf{test} & 0,414 & P \\
&\textbf{perch\'e} & 0,336 & T & & ahah & 0,33 & P \\
&ahahah & 0,306 & T && xd & 0,330 & P \\
& & & & & sì & 0,318 & P\\
\bottomrule
\end{tabular}
\caption{Results of the \emph{Wordify} tool on Personal-ITY.}
\label{wordify}
\end{table}


\section{Conclusions}

We presented Personali-ITY, a novel YouTube-based Corpus for Personality Prediction in Italian. An exploratory empirical investigation on our new corpus
confirms that identifying MBTI personality trait from social media texts is challenging. 
Lexical features perform best, but they tend to be strictly related to the context in which the model is trained and so to overfit.
Concerning our experiments on {\scshape TwiSty}, our model 
outperforms the original {\scshape TwiSty} results \cite{Verhoeven2016TwiStyAM} for Italian and provides a new baseline on this corpus.
Moreover we performed cross-domain experiments between YouTube and Twitter data, achieving scores above the baseline. Performance drops from the in- to the cross-genre setting show limits in portability, and leave a lot of space for  improvements.

In general, better results could, probably, be obtained using more complex models by using neural networks (e.g. LSTM) \cite{mehta2019recent}.
The choice to use a simpler SVM model was driven by its greater understandability and transparency.
These qualities allowed the features analysis of Section \ref{analysis}.
The inherent difficulty of the task itself is confirmed and deserves further investigations, as assigning a definite personality is an extremely subjective and complex task even for humans.
The distant supervision approach remains promising, also applied to YouTube data. Indeed, assigning an `absolute' personality to an individual is difficult. Even if, to some extent, distant supervision is inaccurate and can lead to the creation of corpus containing bias and noise, 
since there is no control on user statements,
we cannot avoid considering that according to the literature also professional psychological tests can lead to inaccurate results.
Moreover, DS, despite its inaccuracy, allows the availability of a large amount of data, necessary for addressing the task by using machine learning approaches.

\section*{Acknowledgments}

The work of Elisa Bassignana was  partially carried out at the University of Groningen within the framework of the Erasmus+ program 2019/20.

\bibliographystyle{coling}
\bibliography{coling2020}

\end{document}